# HKTGNN: Hierarchical Knowledge Transferable Graph Neural Network-based Supply Chain Risk Assessment

Zhanting Zhou*, Kejun Bi†, Yuyanzhen Zhong‡, Chao Tang§, Dongfen Li¶, Shi Ying∥ Ruijin Wang**,
*School of Information Science and Engineering, University of Electronic Science and Technology of China,* Chengdu, China***
*School of Computer Science, Sichuan University,* Chengdu, China†
*Chongqing CEPREI Industrial Technology Research Institute Co., Ltd*, Chongqing, China‡
*Sichuan Changhong Electronic Holding Group Co., Ltd*, Chengdu, China§
*College of Computer Science and Cyber Security, Chengdu University of Technology,* Chengdu, China¶
*School of Computer Science, Wuhan University,* Wuhan, China∥
*Email:* ztzhou@std.uestc.edu.cn*, bikejun@stu.scu.edu.cn†, 1031545418@qq.com‡,
xsb.tangchao@changhong.com§, lidongfen17@cdut.edu.cn¶, yingshi@whu.edu.cn∥, ruijinwang@uestc.edu.cn**

*Abstract*—The strength of a supply chain is an important measure of a country's or region's technical advancement and overall competitiveness. Establishing supply chain risk assessment models for effective management and mitigation of potential risks has become increasingly crucial. As the number of businesses grows, the important relationships become more complicated and difficult to measure. This emphasizes the need of extracting relevant information from graph data. Previously, academics mostly employed knowledge inference to increase the visibility of links between nodes in the supply chain. However, they have not solved the data hunger problem of single node feature characteristics. We propose a hierarchical knowledge transferable graph neural network-based (HKTGNN) supply chain risk assessment model to address these issues. Our approach is based on current graph embedding methods for assessing corporate investment risk assessment. We embed the supply chain network corresponding to individual goods in the supply chain using the graph embedding module, resulting in a directed homogeneous graph with just product nodes. This reduces the complicated supply chain network into a basic product network. It addresses difficulties using the domain difference knowledge transferable module based on centrality, which is presented by the premise that supply chain feature characteristics may be biased in the actual world. Meanwhile, the feature complement and message passing will alleviate the data hunger problem, which is driven by domain differences. Our model outperforms in experiments on a real-world supply chain dataset. We will give an equation to prove that our comparative experiment is both effective and fair.

*Index Terms*—Hierarchical Graph Neural Networks, Risk Assessment, Supply Chain, Graph Data Mining

## I. INTRODUCTION

The complexity of supply chains develops with the number of companies, posing issues in assessing supply chain risk. These problems are accompanied by the digital revolution of traditional corporate processes and supply connections. Investors and suppliers have already recognized that these issues may provide a future competitive edge[1]. Machine learning approaches have been given in recent years to complete link prediction or cluster distinct labels, in addition to early methods based on graph theory and complex networks to construct a risk assessment model of supply chain[3]. These methods are gaining popularity among academics who use knowledge graphs (KGs) or graph neural networks (GNNs) for knowledge inference[4]–[6].

Graph data mining of supply chain networks has some parallels with social networks[6]. Because the topics of the supply chain who will be supervised include individuals and suppliers, the supply chain's complexity and specific suppliers' market objectives make them hesitant to disclose entire data. As a result, the usual data collection method yields partial graph information[4]. Existing work in the machine learning field focuses on link prediction to overcome these difficulties. In addition to using topology-based frameworks[5], knowledge graphs[4], and graph neural networks[6] for link prediction, Brintrup et al. proposed a "Digital Supply Chain Risk Surveilliance" (DSCRS) framework in 2022[4]. These methods use machine learning to mine data, improve supply chain data visibility, and enable relevant professionals to collect helpful information from the supply chain, such as hidden supply relationships and ingredient relationships, to improve supply chain visibility[1]. However, link prediction methods do not fundamentally solve the problem of collecting supply chain data from the source.

Simultaneously, enhanced visibility of supply chain relationships has yet to successfully address the issues of data hunger and distribution shift[2]. On the one hand, due to observational limitations, incomplete field knowledge, and distribution-shift issues[7], [8], the collected graphs frequently contain incomplete node features and unavailable labels, causing data hunger in a variety of machine learning applications[9], [10]. Non-listed companies, on the other hand, are not obligated to release their financial statements, and listed companies

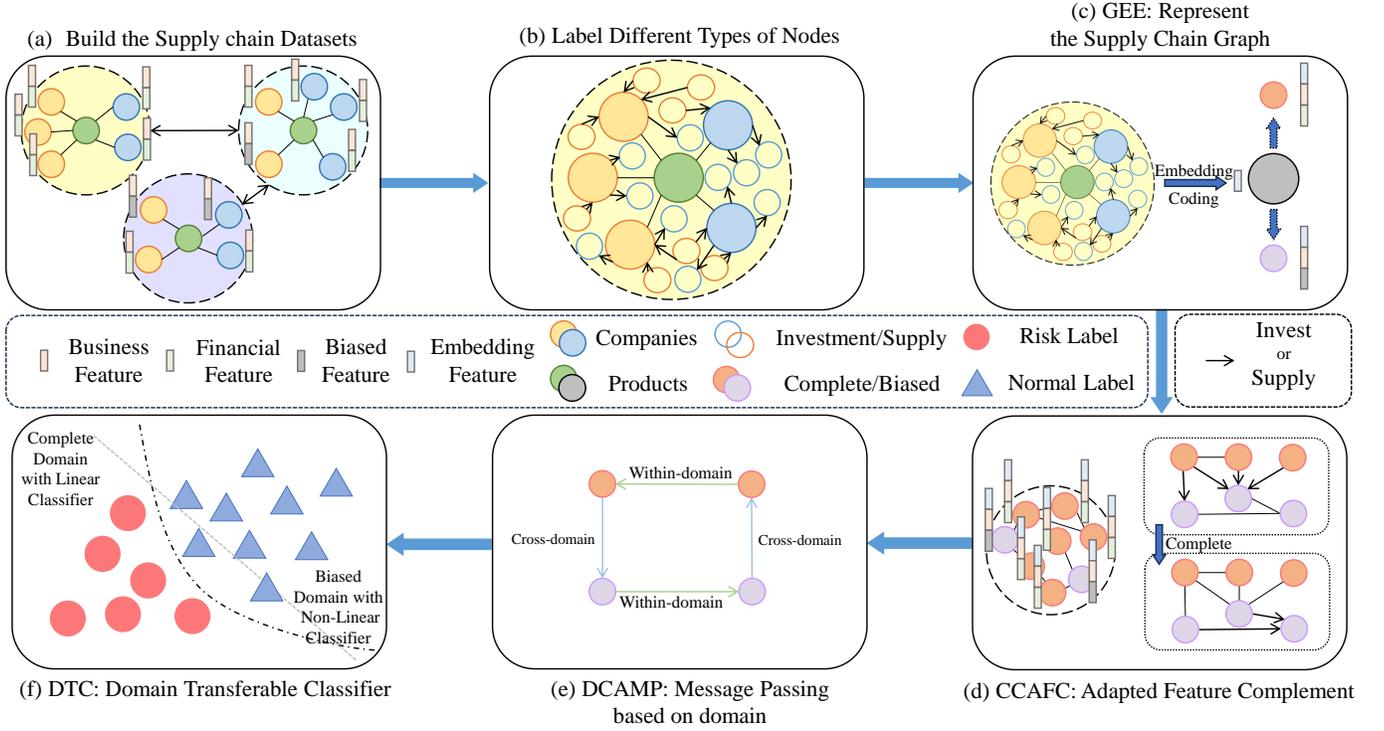

Fig. 1. The overall flow chart of Hierarchical Knowledge Transferable Graph Neural Network-based(HKTGNN) Supply Chain Risk Assessment. (a) and (b) is the data pre-process: (a) shows how to build a supply chain dataset, and (b) shows how to label different types of nodes; (c) show that we use GEE module to embed single-product graph; (d) shows that we use CCAFC module to complete biased nodes; (e) shows that we use DCAMP module to realize domain-difference-based message passing; and (f) shows that we use DTC module from [2] to bi-classify. We will show our model's details in Sec.III

may have missing financial statements in a given quarter, resulting in missing data for some nodes[11]. Both types of companies are important for financial risk assessment[12]. These information is more about the specific information of nodes in the supply chain, and it is not enough to mine graph data only from the perspective of link prediction. Bi et al. proposed a knowledge transferable module is proposed, which can effectively complete biased feature attributes, solve the problem of data hunger, and solve the problem of distribution shift by message passing based on domain differences[2].

The supply chain contains many supply chains for a single product, making the data volume of graph data extensive. The significant number of nodes, edges, and feature attribute dimensions from supply chain networks require and consume a lot of computing resources, and often cause out-of-memory problems[13]. How to reduce the usage of computing resources while retaining as much information as possible in the graph has also become one of the difficulties in practical applications. Representation learning is a commonly used solution that can transform high-dimensional vectors into low-dimensional dense vectors[14]. In the field of investment risk assessment, Bi et al. proposed a module for tribal-style graph embedding that distinguishes between listed companies and non-listed companies, which can also effectively reduce the consumption of single-time computing resources and avoid the out-of-memory problem[11].

We propose a novel hierarchical knowledge transferable graph neural network model (HKTGNN) to address these issues. Considering that only three types of directed link, which named investment-or-supply-to-companies link, companies-product link and product-to-product link, are easier to collect, our modeling will be based on these links, avoiding the problem of it being difficult to collect real company relationships which may be hidden. In the existing graph data modeling, all graph data are often constructed into knowledge graphs, making the related nodes form into the heterogeneous graphs. The performance of previous non-link-prediction algorithms often needs to be improved, and the performance of link-prediction models is affected by their asymmetric feature attributes. In order to reduce the graph scale and make the most of graph data, we use the idea of hierarchy to convert the originally large-scale heterogeneous graph into a smaller product homogeneous graph. Specifically, in order to meet the needs of reality, we construct the supply chain network through investment-or-supply-to-companies link, companies-product link and product-to-product link; for each product in the supply chain, they each constitute a single-product network; a supply chain network is composed of its total number of single-product supply chain networks. Then, we will use the graph representation learning made by the Graph Embedding Encoder to obtain its graph embedding feature attributes. The supply chain network is embedded into the products supply chain network as a node, and its feature attributes are made by graph embedding features, financial information, and financial statements. At the same time, in order to alleviate the problem of data hunger, we use the domain difference knowledge transferable module which based on centrality, and divide the nodes into complete nodes and biased nodes through CB-Graph[1], realizing the feature complementor and within-domain/cross-domain message passing for data-biased nodes. By comparing the performance of the baseline models, we prove that our algorithm performs better and is more suitable for the real-world situation; and show the process of selecting hyper-parameters and the results of the ablation test.

In general, our contributions are as follows:

- We build our model based on investment-or-supply-to-companies links, companies-product links and product-to-product links. These kinds of links are easier to collect, avoiding the challenging problem of collecting genuine company relationships due to reality factors.
- Through the complete and biased nodes in CB-Graph, we effectively distinguish the data-biased nodes. By implementing a feature complement and within-domain/cross-domain message passing through the domain difference knowledge transferable module based on centrality, we alleviate the problem of data hunger and make our model perform better.
- Through the idea of hierarchy, we convert the large-scale heterogeneous graph, of the whole supply chain network into a smaller product-only homogeneous graph, and embed the product-centred supply chain network as a single product node through graph embedding. This hierarchical method enables us to focus on the risk assessment of each product in the supply chain instead of scattered companies. We will give an equation to prove that our comparative experiment, comparing the risk assessment performance between product nodes and company nodes, is both effective and fair.

## II. PRELIMINARY

### A. Graph Neural Networks

Graph Neural Networks(GNNs) are a method of learning graph representations. They are often used in semi-supervised node classification tasks[15], [16].For given graph $\mathcal{G}(V, E, X, Y)$ as an input dataset, where $V$ is the set of nodes in the graph; $E$ is the set of edges in the graph; $X \in \mathbb{R}^{|V| \times |D_{feature}|}$ represents the feature attributes of the nodes; $Y \in \mathbb{R}^{|V| \times 1}$ represents the labels of the nodes used for classification, The current mainstream graph neural networks use multi-layer graph convolution layers and message-passing aggregation to learn the representations of nodes. The updated formula for each layer is as follows:

$$h_{(l)} = \mathbf{U}(h_i^{(l-1)}, \mathbf{M}(h_i^{(l-1)}, h_j^{(l-1)} | v_j \in \mathcal{N}(v_i))) \quad (1)$$

where $h_i^{(l)}$ represents the node representation learning vector of the $l-th$ layer of GNNs; $\mathbf{M}$ represents the message equation

---

[1]The definition of CB-Graph will be given in Sec. II.C

for aggregating neighbor features; $\mathbf{U}$ represents the update equation.

### B. Complete Nodes&Biased Nodes

We need to model the supply chain graph data we obtain for the supply chain risk assessment task. These graph data have complex and diverse nodes (types) and edges (types), and their feature attributes are essential for representation learning. However, in the real world, there are often problems in collecting their feature attributes, such as completely or partially missing financial statements that cannot be obtained. This requires us to define a new label to help us train the model. We call the nodes with complete data as complete nodes and those with biased data as biased nodes. In addition to the cases where non-listed companies do not have financial information and listed companies do not discover their financial statements, there are also some cases where the embedding vectors are biased due to structural missing in the construction of supply chain graph data, for examples, some products lack companies-product link or we cannot collect certain investment-or-supply-to-companies link from our sources. This is called integrity missing in supply chain risk assessment, and we will label it as a biased node, assign it a value of 0 and label it as a risk. At the same time, for the case of nodes in representation learning with companies with missing business information, we will also label this type of node as biased.

### C. CB-Graph

CB-Graph, denoted by $\mathcal{G}^{CB}(V, E, X, Y, \Psi)$, is a set of graph data used for HKTGNN training in this paper, where $V$ represents the set of nodes in the graph, including complete nodes and biased nodes, with $v_{ij} \in V = V^{complete} \cup V^{biased}$. $E$ represents the set of edges in the graph. $X \in \mathbb{R}^{|V| \times |N_1+N_2+N_3|}$ represents the feature attributes of all nodes in the graph, i.e., $X_i = [X_i^E \parallel X_i^B \parallel X_i^F] = [X_i^o \parallel X_i^u]$, where $X_i^o = [X_i^E \parallel X_i^B]$ represents observable feature attributes, and $X_i^u = X_i^F$ represents unobservable feature attributes that need to be completed. $Y$ represents the risk label, which is marked as having risk when there is a risky company in node $i$. $\Psi$ is the CB-Graph label, which marks which nodes in the graph are complete nodes and which are biased nodes, i.e.:

$$\psi = \begin{cases} 1, & \text{if } v_{ij} \text{ is a biased node} \\ 0, & \text{if } v_{ij} \text{ is a complete node} \end{cases} \quad (2)$$

## III. METHODOLOGY

In this chapter, we will introduce our model in detail. We will first describe the algorithm flow in general, and then introduce the following modules separately: (1) Data Preprocessing Methods; (2) Graph Embedding Encoder (GEE); (3) Cross-domain Centrality-based Adapted Feature Complementor (CCAFC); (4) Domain Centrality-based Adapted Message Passing module (DCAMP).

## A. Model flow

Our model flow is shown in the following flow chart 1. First, we pre-process the data to obtain $\mathcal{G}_{supply}^{CB}$; Then, we use the GEE module to obtain $\mathcal{G}_{product}^{CB}$; After randomly selecting a random seed, we randomly split the dataset. After the convolution layer, we use the CCAFC module to complete feature attributes. We use the DCAFC module to implement within-domain and cross-domain message passing. Finally, we use the DTC algorithm in [2] to perform bi-classification for different domains to obtain the score.

**Algorithm 1** HKTGNN

1: **Input**: $D_s$, model parameters $\omega$; hyper-parameter $\lambda$, $\gamma$, $K$, $Dim_e$, $n$, $m$, $seeds$.
2: **Output**: $Y^p$
3: **function** HKTGNN($D_s$)
4:     $D_s \to \mathcal{G}_s^{CB}$ with $ndata$
5:     $X_E \in \mathbb{R}^{|V_P| \times Dim_e} \leftarrow GEE(\mathcal{G}_s^{CB})$
6:     $X \leftarrow [X_E \parallel X_B \parallel X_F] \in \mathcal{G}_p^{CB}$
7:     **for** $i = 1$ to $n$ **do**
8:       Randomly split $\mathcal{G}_p^{CB}$ with a specific segment proportion
9:       $x_{conv} \leftarrow U_{conv}(\mathcal{G}_p^{CB})$
10:       $x_{cple} \leftarrow CCAFC(x_{conv})$ with Eq.5-9
11:       **for** $j = 1$ to $m$ **do**
12:         $x_m \leftarrow DCAMP(x_{cple})$ with Eq.10-14
13:       $Y^{pre} \leftarrow DTC(x_m)$
14: **function** GEE($\mathcal{G}_s^{CB}$)
15:     Initialize $ndatas$ of $\mathcal{G}_s^{CB}$
16:     Embedding($ndatas$)
17:     return $X_e$
18: **function** CCAFC($x^o$)
19:     Initialize $\omega^c$, $\omega^b$
20:     **for** $k = 1$ to $p$ **do**
21:       $x_{cal} = \sum x_{eli} f(x_i^o \omega^b, x_j^o \omega^c)$ as Eq.5
22:     return $\omega^c$, $\omega^b$
23: **function** DCAMP($x_m$)
24:     Initialize $h_i \leftarrow x_m$
25:     **for** $l = 1$ to $q$ **do**
26:       $\mathcal{M}_{v_i \to v_j} = \tilde{h}_i \cdot f^{\Psi(v_j)}(\tilde{h}_i, h_j)$ as Eq.10
27:       $\triangle X^C \leftarrow h_i$ as Eq.14
28:     $w^{fin} \leftarrow h_i$
29:     return $w^{fin}$

## B. Data Processing Method

In previous GNNs-based data mining methods, it is generally to construct an entire large graph with all company nodes and product nodes. Considering that the investment-or-supply-to-companies link, product-to-product link and companies-product link with related financial statements are easier to obtain, different types of nodes such as products, companies, investments, and suppliers are present in the node set, and different types of investment relationships and supply relationships are present in the edge set. Finally, it will be constructed by building a directed heterogeneous graph with a large number of nodes. Through hierarchical processing and graph embedding, we will build a directed homo-generous graph with fewer nodes and Start message passing. Here, we define $\mathcal{G}^{supply}$ as the heterogeneous graph of entire supply chain networks; $\mathcal{G}^{single\_product\_supply}$ as the heterogeneous graph of supply chain networks which only built about a single product; and $\mathcal{G}^{product}$ as the only product network homogeneous graph used as the bottom layer of the hierarchy for training. Figure 2 shows a schematic diagram of the hierarchical idea. In our work, the following problems may exist in the processing of directed heterogeneous graphs: since a company may exist in multiple supply chains at the same time, its business information and financial statement may be utilized multiple times; at the same time, since we are doing representation learning on the single-product supply chain network, the topology features such as the degree of company nodes in the directed heterogeneous graph may change. In order to solve this problem, we will use the data preprocessing method of business information and the design of the encoder for representation learning to compensate for these problems. Our experimental results show that our processing is effective.

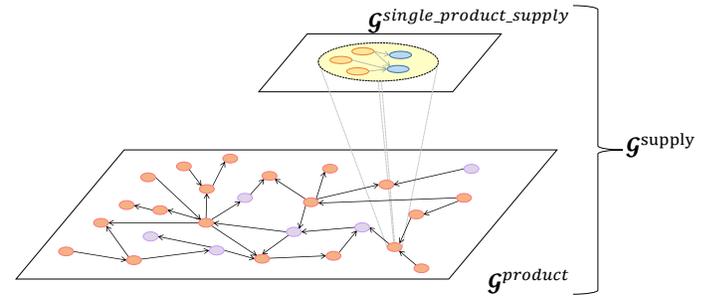

Fig. 2. Image of A Hierarchical Graph Structure for Supply Chain Risk Assessment

We constructed a real-world dataset based on the data provided by Sichuan Changhong Electric Appliances Co., Ltd.[2] on the integrated circuit supply chain, the financial statements of enterprises in the integrated circuit supply chain obtained from Wind[3] and RESSET[4], and the investment or supplier information related to the companies of the integrated circuit supply chain in Tianyancha[5]. Specifically, after the hierarchical and graph embedding, for the product CB-Graph CB-Graph$\mathcal{G}_{product}^{CB}(V, E, X, Y, \Psi)$ given to us, the represen-tation learning features $X^E$ are from the GEE of the hierarchical structure, with a dimension of $\mathbb{R}^{N_1=64}$; the financial information features $X^B$ come from the z-score normalization of Changhong's data and the one-hot encoding of product business information, with a dimension of $\mathbb{R}^{N_2=17}$; the financial statement features $X^F$ are the z-score normalized results of the financial statements published in Wind and RESSET, with a dimension of $\mathbb{R}^{N_3=18}$. For each representation learning node i, we take the financial statements of each product node $X_i^F$ as follows: $X_i^F = \overline{X_i^F}$, where $\overline{X_i^F}$ represents the mean of the financial statements of all companies in node $i$. We use the stock rating from Tongdaxin[6] to define the risk label of listed enterprises: 0-80 points are set as medium risk or high risk; 80-100 points are set as low risk or no risk; for a product, the risk of its supply chain network is determined by the risk labels of the companies in it. For the CB labels of nodes, we mainly define the following situations as biased: (1) missing investment or supply relationships about non-listed companies or natural persons; (2) missing or incomplete financial statements of listed companies; (3) unable to obtain the relevant supply chain companies for the corresponding product. When a company in a product has any of the (1) or (2) situations, it is marked as biased. Therefore, it can be seen that it is not necessary for the numerical values of feature attributes of biased product nodes to be process all 0.

---

[2]Changhong is a Chinese multinational consumer electronics company and our program collaborator: https://cn.changhong.com/
[3]Wind is a leading financial data and information provider in China: https://www.wind.com.cn/
[4]RESSET is a leading financial data and research software developer in China: http://www.resset.com/
[5]Tianyancha is a Chinese company information platform that provides comprehensive information on Chinese companies: https://tianyancha.com

### C. Graph Embedding Encoder(GEE)

After the construction of each single-product supply chain network, we need to start representation learning and embed them into the supply chain network through an encoder to form a new homogeneous directed graph that is easy to train: each node represents a single-product supply chain network and carries its feature information; each edge represents a product-to-product link, connecting the nodes. We designed the GEE module to embed a single-product supply chain network to achieve this goal. Traditionally, topological information such as in-degree, out-degree, shortest path, eigenvalues, and node types can be used to measure the characteristics of the graph[17]–[19]. Due to the specific background settings, we design our GEE as follow. The general flow of GEE is shown in Fig. 1(c).

Our graph embedding goal is to embed the single-product supply chain network information as the feature attributes of the nodes in product network. For the $i-th$ single-product supply chain graph $g_i^{single\_product\_supply}$, with node set $V_i^{supply}$, we take $v_j^{supply} \in V_i^{supply}$ for each unattributed node, we mainly consider the following topological information:

1) The degree of each node. We denote the in-degree of each node as $deg_j^+$ and the out-degree as $deg_j^-$.
2) The node type of each node $\phi_j$ Here, we have four types of nodes: (1) product nodes; (2) listed company nodes in the single-product supply chain; (3) investment nodes of listed companies (including enterprises and individuals); (4) invested nodes of listed companies (including enterprises and legal persons).
3) The shortest path of each node $SPB_j$. Since we have reduced the scale of the graph, the time cost and memory cost of the shortest path is greatly reduced.

We use the above information to obtain the graph embedding layer and its embedding information:

$$Emb(v_j^{supply}) = EmbeddingLayer(deg_j^+, deg_j^-, \phi_j, SPB_j) \quad (3)$$

In addition, as mentioned in [20], [21], we introduce the top Laplacian vector of each node. For each node $v_j^{supply} \in V_i^{supply}$, the maximum eigenvalue of the node $v_j^{supply}$ is the $j-th$ value of the top Laplacian vector. Finally, we get the feature value of each node:

$$Z(g_j^{supply}) = \left[ Emb(v_j^{supply}) \parallel u_o(g_j^{supply}) \right] \quad (4)$$

where $u_o(g_j^{supply})$ is the top Laplacian vector of $g_j^{supply}$. After we get $Z(g_j^{supply})$, we put the supply chain network and the vector of each node into the Graph Isomorphism Network (GIN)[22] for training, and finally get the embedding vector.

### D. Cross-domain Centrality-based Adapted Feature Complementor(CCAFC)

After using the GEE module, we get a graph with product-to-product links as edges and a single-product supply chain network as nodes. They have the embedding features obtained in GEE, business information feature attributes and financial statement feature attributes mentioned in Sec.III.B. At this time, we get the CB-Graph mentioned in Sec.II.C. To complete the features of all biased nodes, we propose the CCAFC module based on the literature [2]. In order to explain our cross-domain feature complement and lay the foundation for the message passing module later, we first supplement the module's settings on the CB-Graph here. We first stipulate that $V^{complete} \cup V^{biased} = V^{supply}$. At the beginning of the module execution, we have $\mathcal{V}^+ = V^{complete}$ and $\mathcal{V}^- = V^{biased}$; as the knowledge in $\mathcal{V}^+$ is transferred to $\mathcal{V}^-$ to complete feature complement, $\mathcal{V}^-$ will become $\mathcal{V}^+$, with $V^{complete} \subset \mathcal{V}^+ \subset V^{supply}$. We use a hyperparameter $K$ to limit the number of iterations of each transformation. This is the general flow of the CCAFC module which is shown in Fig. 1(d). For each uncompleted node $v_i \in \{V^{biased} \cap \mathcal{V}^-\}$, its unobservable vector $x_i^u$, after completing the complement of CCAFC, is:

$$\widehat{x_i^u} = \sum_{v_j \in \{\mathcal{N}(v_i) \cap V^{complete}\}} \tilde{x}_j^u \cdot f(x_i^o \omega^b, x_j^o \omega^c) \quad (5)$$

where $\widehat{x_j^u}$ is the calibrated variable for $x_j^u$ which eliminate the domain differences:

$$\begin{cases} \tilde{x}_j^u = x_j^u - \triangle \tilde{X}^u \cdot \sigma([x_j^u \parallel \triangle \tilde{X}^u] \cdot \omega^g) \\ \triangle \tilde{X}^u = \\ \left[ \sum_{i=1}^{|V^{complete}|} x_i^{oc} \cdot C_i^c - \sum_{j=1}^{|V^{biased}|} x_j^{ob} \cdot C_j^b \right] \cdot \omega^{o \to u} \end{cases} \quad (6)$$

---
[6]Tongdaxin is a leading Chinese stock trading software developer: https://www.tdx.com.cn/

where $\omega^g \in \mathbb{R}^{2D^u \times D^u}$ and $\omega^{o \to u} \in \mathbb{R}^{D^o \times D^u}$ are self-learned parameter matrices; $\sigma$ is the softsign activation function; $\triangle \tilde{X}^u$ is the transform domain difference based on centrality, where $x_i^{ov}$ is the attribute of the $i$th complete node, and $x_j^{os}$ is the attribute of the $j$th biased node; Considering that we use graph embedding vectors as visible feature attributes, which only contain topological information in a single-product supply chain network, we make the algorithm better by using centrality weighting based on the idea of weighted average. Among them, $C_i^v$ is the centrality of the $i$th complete node, and $C_j^s$ is the centrality of the $j$th biased node. Specifically, the feature vector centrality[23], [24] is used as the calculation method in the background of this paper. As shown in Figure 1(d), after a biased node is completed, the next iteration will use the supplemented biased node as the information provider, realizing the transformation from $V^- \to V^+$. In this process, we use breadth-first search (BFS), with eqn.7:

$$\widehat{x_i^u} = \sum_{v_j \in \{\mathcal{N}(v_i) \cap V^+\}} x_j^u \cdot f(x_i^o \omega^c, x_j^o \omega^b) \quad (7)$$

where $f(\cdot)$ is the neighbor importance factor:

$$f(x_i^o W^c, x_j^o W^b) = \sigma([x_i^o W^c \parallel x_j^o W^b] \cdot a^{cb}) \quad (8)$$

where $\sigma$ is the PReLU activation function; $a^{cb}$ is a calibrated weight computed by each $x_i^o$ and $\mathcal{N}(v_i)$. Through iteration, the algorithm converges when the number of iterations reaches the maximum value $K$ or all nodes are fully completed. In order to ensure that $\triangle \tilde{X}^u$ in Eq. 9 is fully applied, we also introduce the distribution consistency loss $\mathcal{L}_c^{dist}$ from [2] in the iterative process, and make changes about centrality:

$$\mathcal{L}_c^{dist} = \| \triangle \tilde{X}^u - \left( \sum_{i=1}^{|V^{complete}|} x_i^u \cdot C_i^c - \sum_{i=1}^{|V^{biased}|} \widehat{x_i^u} \cdot C_i^b \right) \|_2^2 \quad (9)$$

In Sec.IV, we will demonstrate that our changes perform better on supply chain network data than the original formula.

### E. Domain Centrality-based Adapted Message Passing modules(DCAMP)

After completing the feature, we need to pass the message on the homogeneous graph with attributes and finally complete bi-classification. Here we introduce the message passing on CB-Graph, as shown in Figure 1(e). Specifically, there are four cases: (1) message passing between complete nodes and biased nodes; (2) message passing between biased nodes and biased nodes; (3) message passing between biased nodes and complete nodes; (4) message passing between complete nodes and complete nodes. Among them, (1) and (3) are out-of-distribution (OOD) cross-domain message passing; (2) and (4) are in-distribution (ID) intra-domain message passing. Cross-domain OOD message passing can be noise that lowers model performance[2], [7], [25], so the DCAMP module is needed to solve this problem: by calculating the domain difference ratio factor, the OOD feature of the source node is projected to the target node domain to transmit messages. We also refer to the practice in [2], and make some improvements.

DAMP mainly considers two factors: (1) bidirectional domain difference based on centrality; (2) neighbor importance. For a given edge $e_{i,j} = (v_i, v_j)$, the message passing equation is as follows:

$$\mathcal{M}_{v_i \to v_j} = \tilde{h}_i \cdot f^{\Psi(v_j)}(\tilde{h}_i, h_j) \quad (10)$$

where $\tilde{h}_i$ is the feature calibrated by the domain difference factor; $f^{\Psi(v_j)}(\cdot)$ is the neighbor importance function:

$$f^{\Psi(v_j)} = \sigma([\tilde{h}_i \cdot \omega^{\psi(v_j)} \parallel h_j \cdot \psi(v_j)]) \cdot a^{\psi(v_j)} \quad (11)$$

$$\tilde{h}_i = \mathcal{P}_{\psi(v_i) \to \psi(v_j)}(h_i) = h_i + \triangle_{i,j} \quad (12)$$

where $\sigma$ is the $leaky\_ReLU$ activation function; $a^{\psi(v_j)}$ is a GATv2[26] attention weight; $\triangle_{i,j}$ is the distribution-shift variable:

$$\triangle_{i,j} = \begin{cases} 0, & \text{if} \psi(v_i) = \psi(v_j) \\ (-1)^{\psi(v_j)} \cdot \\ \quad \sigma([h_i \parallel \triangle X^C] \cdot a^{\psi(v_j)}) \cdot (\triangle X^C), & \text{else} \end{cases} \quad (13)$$

where $\sigma$ is the softsign activation function; $\psi(v_i)$ is the label; as Eq. 14, with $\triangle X^C$:

$$\triangle X^C = \sum_{i=1}^{|V^{complete}|} x_i^{oc} \cdot C_i^v - \sum_{j=1}^{|V^{biased}|} x_j^{ob} \cdot C_j^s \quad (14)$$

We used centrality-based domain difference to realize cross-domain message passing. After processing through our modules, we use DTC module from [2] to realize bi-classification, and our $\mathcal{L}^{total}$ as follow:

$$\mathcal{L}^{total} = \mathcal{L}^{clf} + \lambda \cdot \mathcal{L}^{kl} + \gamma \cdot \mathcal{L}_c^{dist} \quad (15)$$

We will test hyper-parameters $\lambda$ and $\gamma$ widely with analysis in our supply chain dataset in Sec.IV.B.(2).

## IV. EXPERIMENTS

In this chapter, we will introduce our experimental setup background and experimental results and analysis separately.

### A. Experiments Set

*1) Dataset:* We will explain all the graph data we collected at first. In the integrated circuit supply chain, we obtain 430 products, 1,732 listed companies, and 46,273 relevant investment/supply enterprises or natural person nodes, collected by Changhong which is introduced in footnote 2. There are 1,875 product-to-product links and 0 to 3,137 company-product links in a single-product supply chain network, with 53,438 links in the entire supply chain network. The adjacency matrix formed by the graph data is very sparse.

The dataset used by HKTGNN is made in the way described in Sec.III.B. After graph embedding, we finally got a homogeneous graph with 430 product nodes, 1,875 edges, each node with a 99-dimensional feature attribute vector, and marked by risk labels and CB labels.

In contrast to the dataset used in our hierarchical model, we need to adjust it to make it compatible with the baselines. As shown in Tab. I, we use four types of datasets to test our baseline. We show betweenness centrality, degree centrality and power-law centrality of each dataset to demonstrate the effectiveness of our model comparison experiments. The main work on datasets preparation is as follows:

- Because our task is to develop a model that can assess the risks of each link in the supply chain, our hierarchical model focuses on the risks of each products supply chain. However, the product nodes in the supply chain network constructed without using the hierarchical model lack financial data. They cannot be used effectively in the use of homogeneous graphs. Therefore, we follow the approach in [2] and set the nodes to listed companies for the baseline and other models used in the experiment. It is important to note that since a product's supply chain includes multiple listed companies, the risks of each listed company will directly affect the risk of the product's supply chain. At the same time, the definition of CB-Graph is more relaxed than that of the hierarchical model, only considering whether it lacks financial data. It would be feasible to argue that even if the baseline and other models achieve the same score as our model, their actual supply chain risk assessment will still be poor because of scattered companies in the entire supply chain. Specifically, considering each kind of accuracy, $acc_{ours}$ of each product-only supply chain and $acc_{company}$ of each company in the certain product-only supply chain, we can proof that when considering the risk assessment of the product-only supply chain, which can represent the risk of whole supply chain, there is an inequality:

$$acc_{ours} \geqslant \frac{\sum_{k=1}^{N}(acc_{company}^{k} \times R_k)}{N} \qquad (16)$$

  where $N$ is the number of companies in the certain product-only supply chain; $R_k \in [0,1]$ is the risk-assessing variable for each company, represents the risk of companies' position feature and other business features which have been embedded by GEE module in our method. This is why our comparative experiment is both effective and fair.

- Since we do not have direct connection between compaines in the supply chain network, we only made two types of datasets: (1) when the product-to-product link exists, all listed companies in A product will be fully connected to all listed companies in B product or all listed companies in A will point to all listed companies in B. (2) When the product-to-product link exists, the listed companies in A product will be randomly connected to the listed companies in B product with a certain random degree. For the each random-join graph in (2), we will generate ten instances, and the final comparison results will be taken as their average, as shown in Table I.

- The risk labels are made in the same way as the risk labels of our proposed model. For the feature attributes of each node, the financial data are from the financial statements of each listed company. However, the business information comes from the average financial information of all products related to the listed company.

*2) Baselines:* In comparative experiment, we mainly used the following models for comparative experiments:(1) MLP[27], a neural network of multiple hidden layers. (2) GGNN[28], a graph neural network that uses gated recurrent units and modern optimization techniques, and then extends to the output sequence.(3) GraphSage[29] and (4) GraphSage2[29], both are graph neural network models based on neighbor aggregation, used for graph node representation learning.(5)GATv2[26] is an enhanced version of GAT, a well-known graph attention network. (6) KTGNN[2] is a knowledge-transferable graph neural network that can cope with data hunger and distribution drift. (7)HKTGNN(*ours*).

*3) Training Details:* In our tests, we found that for KT-GNN, we set hyper-parameters as [2] which will perform the best. All other baseline have been adjusted to the best performance on the supply chain dataset. For our model HKTGNN, we set $\lambda$(the weight of KL-divergence loss in DTC module) to 0, $\gamma$(weight for the centrality-based distribution consistency loss term $l_c^{dist}$) to 0.1, and $K$(steps of feature completion in CCAFC module) to 2. We set the parameters of our modules as follow: the learning rate $lr$ is set to $3e-2$, the data set split ratio is $7:1:2$, and it is randomly divided. We will display our reasons for choosing these parameters and hyper-parameters, and give the analysis of our model and dataset, please refer to Sec.II.B.2. The remaining hyper-parameters are the same as [2]. We conducted ten times Monte Carlo experiments [30] for each dataset experiment, each with 300 iterations. Our tests were conducted on NVIDIA GeForce RTX 2070 SUPER hardware and Python3.10 as the experimental environment.

### B. Results

We will present our experimental results in two aspects: comparing scores in comparative experiment and analyze our model in further tests. We selected the F1 score and AUC score, commonly used in binary classification, as evaluation indicators. F1 score measures the balance between precision and recall and AUC score measures the ability of a model to distinguish between positive and negative classes. All of our experiments use these evaluation metrics to compare the performance of different condition. For our model scores, we are going to bold them; For the second one in the comparison score, we will underline it.

*1) Comparative Experiment:* As can be seen from Table II, our model HKTGNN has a significant improvement in both F1 score and AUC score on the real supply chain dataset, compared with the well-known baseline models and the relatively new financial risk assessment model KTGNN. The datasets displayed in Table I and analysis in secIV.A show the completeness and rigorous of our comparative experiment. Considering the proof as eqn.16, the perform in comparative experiment indicates that our model is better in the real scenario and can effectively solve the data-hunger problem in supply chain risk assessment. Specifically, our model performs better on the bi-classification task and for supply chain risk assessment models; directly assessing the risk of a product-centred supply chain network is more effective than assessing the risk of companies scattered in the supply chain network. This is because the risk of a single-product supply chain network is determined by the risks of all the companies in the single-product network. By directly assessing the risk of the entire network, we can get a more comprehensive and accurate risk assessment.

TABLE I
DATASETS FOR BASELINE. WE DIVIDED THE DATASETS INTO FOUR GROUPS: (1) FJG: FULLY CONNECTED GRAPH; (2) 25%G: GRAPH WITH 25%
RANDOM CONNECTIVITY; (3) 50%G: GRAPH WITH 50% RANDOM CONNECTIVITY; (4) 75%G: GRAPH WITH 75% RANDOM CONNECTIVITY. THE SPECIFIC
WAY TO GENERATE CONNECTIONS IS PROVIDED IN THE DATA GENERATION SECTION BELOW. WE COMPARED BETWEENNESS CENTRALITY, DEGREE
CENTRALITY, AND POWER-LAW CENTRALITY TO SHOW THAT OUR RANDOM DATASETS COVER A WIDE RANGE OF DEGREES AND SHORTEST PATHS,
WHICH CAN BETTER DEMONSTRATE THE EFFECTIVENESS OF OUR MODEL COMPARISON EXPERIMENTS. SINCE THE RANDOM CONNECTIONS MAY CAUSE
SOME NODES TO BECOME ISOLATED POINTS, WE WILL REMOVE THEM TO AVOID THE IMPACT ON BASELINE MESSAGE PASSING.

| Name of Datasets | Quantity of Edges | (%)Average Betweenness Centrality | (%)Average Degree Centrality | (%)Average Power-law Centrality | (%)Average Closeness Centrality |
|---|---|---|---|---|---|
| $FJG$ | 277237 | 0.0662301 | 18.8193480 | 0.0582411 | 35.8250008 |
| $25\%G[0,9]$ | [35397, 35896] | [0.0774914, 0.0830326] | [2.4843665, 2.5151257] | [0.0589971, 0.0593120] | [17.3716286, 17.5991133] |
| $50\%G[0,9]$ | [71305, 71914] | [0.1810510, 0.1840582] | [17.6055292, 17.8410984] | [0.0809061, 0.0814996] | [31.3224846, 31.3981867] |
| $75\%G[0,9]$ | [106748, 107351] | [0.0599717, 0.0611433] | [7.3340517, 7.3894729] | [0.0586166, 0.0586854] | [23.1434710, 23.2686373] |

TABLE II
TEST RESULTS FOR EACH BASELINE, WITH F1 SCORE AND AUC SCORE. WE COMPARED A TOTAL OF FIVE DATASETS MENTIONED IN THE PREVIOUS
TEXT: (1) THE SUPPLY CHAIN DATASET USED BY HKTGNN; (2) THE FULLY CONNECTED GRAPH DATASET USED BY THE BASELINE; (3) THE GRAPH
DATASET GENERATED WITH 25% RANDOMNESS USED BY THE BASELINE; (3) THE GRAPH DATASET GENERATED WITH 50% RANDOMNESS USED BY THE
BASELINE; (3) THE GRAPH DATASET GENERATED WITH 75% RANDOMNESS USED BY THE BASELINE. THE SCORES OF (2), (3), AND (4) DATASETS ARE
THE AVERAGE OF TEN DATASETS.

| Model Name | Full Join Graph(2) or Complete Graph(1) | | 25% Random Join Graph(3) | | 50% Random Join Graph(4) | | 75% Random Join Graph(5) | |
|---|---|---|---|---|---|---|---|---|
| | F1 | AUC | F1 | AUC | F1 | AUC | F1 | AUC |
| MLP | 64.67 ± 0.04 | 62.86 ± 0.03 | 67.31 ± 0.04 | 65.19 ± 0.03 | 65.68 ± 0.01 | 63.73 ± 0.01 | 67.58 ± 0.04 | 65.43 ± 0.03 |
| GGNN | 62.22 ± 0.04 | 61.01 ± 0.04 | 62.80 ± 0.04 | 61.86 ± 0.04 | 61.67 ± 0.05 | 60.73 ± 0.04 | 63.70 ± 0.05 | 62.53 ± 0.04 |
| GraphSage | 67.71 ± 0.01 | 65.75 ± 0.01 | 63.77 ± 0.04 | 62.84 ± 0.02 | 63.41 ± 0.02 | 61.92 ± 0.02 | 64.49 ± 0.05 | 63.76 ± 0.02 |
| GraphSage2 | 67.41 ± 0.02 | 65.42 ± 0.02 | 67.80 ± 0.01 | 66.05 ± 0.01 | 65.51 ± 0.01 | 64.00 ± 0.01 | 67.79 ± 0.02 | 66.14 ± 0.02 |
| GATv2 | 72.30 ± 3.27 | 67.35 ± 2.35 | 75.89 ± 2.84 | 70.83 ± 2.16 | 81.81 ± 0.90 | 76.04 ± 0.77 | 79.13 ± 0.90 | 73.47 ± 0.76 |
| KTGNN | 74.47 ± 0.00 | 55.31 ± 0.88 | 72.39 ± 1.45 | 49.12 ± 1.56 | 73.21 ± 0.80 | 59.80 ± 1.83 | 71.07 ± 1.68 | 55.03 ± 2.94 |
| **HKTGNN** | **83.13 ± 2.76** | **88.41 ± 1.25** | — | — | — | — | — | — |

TABLE III
WE WILL COMPARE THE PERFORMANCE OF OUR MODEL ON DIFFERENT
SEGMENT PROPORTION. WE TEST EXTENSIVELY AND CHOSE SOME OF
THE MOST USEFUL FINDINGS.

| Segment Proportion | F1 | AUC |
|---|---|---|
| 5 : 1 : 4 | 69.93 ± 3.68 | 76.32 ± 2.72 |
| 5 : 4 : 1 | 62.65 ± 1.35 | 67.58 ± 1.87 |
| 5 : 2 : 3 | 70.40 ± 3.36 | 75.82 ± 3.21 |
| 5 : 3 : 2 | 73.38 ± 3.51 | 81.35 ± 2.01 |
| 6 : 2 : 2 | 80.45 ± 3.09 | 87.17 ± 1.80 |
| 6 : 1 : 3 | 80.50 ± 3.54 | 85.70 ± 3.63 |
| 6 : 3 : 1 | 73.82 ± 3.07 | 80.39 ± 3.55 |
| 7 : 2 : 1 | 73.94 ± 3.88 | 78.74 ± 3.66 |
| 8 : 1 : 1 | 78.53 ± 3.49 | 81.50 ± 3.19 |
| 7 : 1 : 2 (**ours**) | **83.13 ± 2.76** | **88.41 ± 1.25** |

TABLE IV
ABLATION STUDY TO DEMONSTRATE THE IMPROVEMENT OF OUR
APPROACH, OF CHANGING THE ALGORITHM WITH CENTRALITY IN
HKTGNN.

| Model Name | F1 | AUC |
|---|---|---|
| HKTGNN(Without Changed) | 75.52 ± 1.19 | 76.30 ± 1.76 |
| **HKTGNN (ours)** | **83.13 ± 2.76** | **88.41 ± 1.25** |

*2) Further Tests:* In this subsection, we will display our further tests to answer three questions: (1)why do we choose 7 : 1 : 2 as our dataset **segment** proportion? (2)why do we choose **hyper-parameters** as above? (3)Do our changes on algorithms with centrality **do** work?

- Dataset Segment Test
  In this part, we mainly compare the impact of our data split on our model's performance, where parameters and hyper-parameters set as same as comparative experiment. We test extensively and chose some of the most useful findings. As shown in Table III, it can be intuitively seen that when we choose the data split as **7 : 1 : 2**, the average F1 score and auc score is the highest. You can find that the F1 score is 3 points higher than the second one 6 : 1 : 3 and nearly 20 points higher than the last one 5 : 4 : 1; the AUC score is 1 point higher than the second one 6 : 2 : 2 and over 20 points higher than the last one 5 : 4 : 1. It shows that dataset segment proportion is crucial for our model. Meanwhile, there are two apparent trends: (1)If we observe with the train set as axis, the test performance shows an obvious trend of decreasing from the highest point to both sides.; (2)The segment as 17 will make our model perform better on our real-world dataset:

$$N_{train} \gg N_{test} > N_{val} \qquad (17)$$

where $N_{sth.}$ presents the number of something.

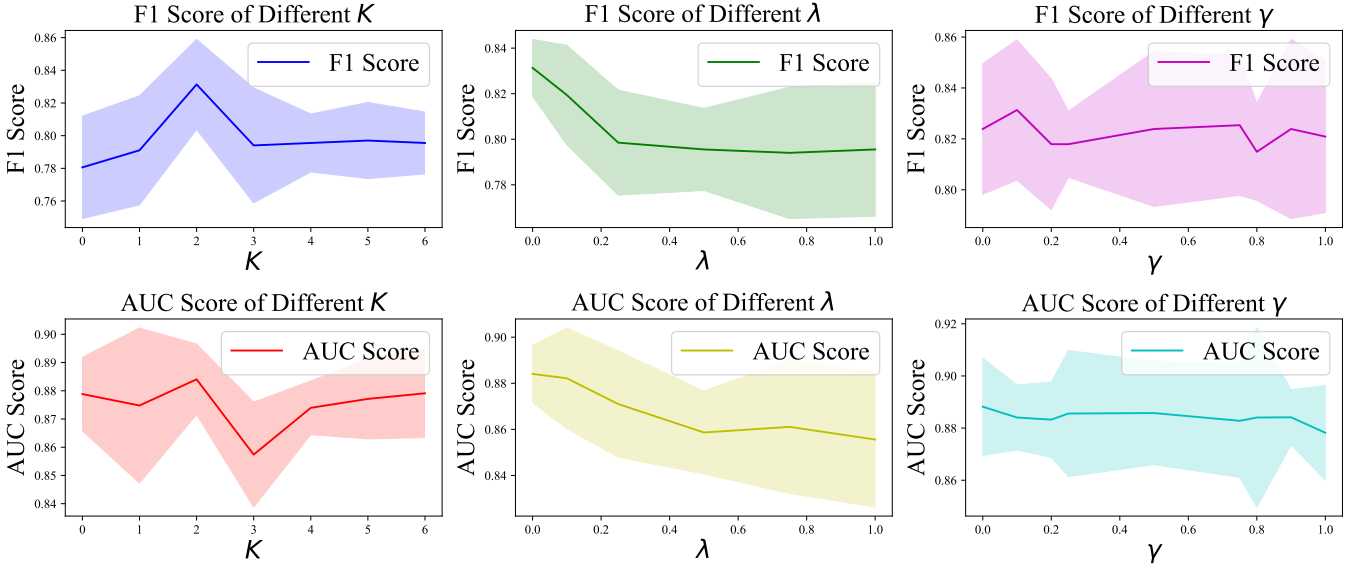

Fig. 3. This chart shows the influence of different $K\backslash\lambda\backslash\gamma$ in F1 Score and AUC Score. (a) & (d) shows the influence of different $K$; (b) & (e) shows the influence of different $\lambda$; (c) & (f) shows the influence of different $\gamma$.

- HyperParameter Test
  1) Test for $K$
     In this part, we test for the hyper-parameter $K \in \{0, 1, 2, 3, 4, 5, 6\}$ as Fig.3(a) & (d), which means the steps of feature completion in CCAFC module. The test performance shows an obvious trend of decreasing from the highest point $K = 2$ to both sides and F1 score always remains at the same level. Meanwhile, setting $K = 2$ makes the performance of F1 score significantly better than $K = 0$, which means there are none of CCAFC module. Accordingly, biased nodes with unobservable feature influence the performance of risk assessment significantly and our CCAFC module works well. We do alleviate the problem of data hunger.
  2) Test for $\lambda$
     In this part, we test for the hyper-parameter $\lambda \in [0, 1]$ as Fig.3(b) & (e), which means the weight of KL-divergence loss in DTC module in 15 influencing the $\mathcal{L}^{total}$. You will find that when $\lambda$ getting smaller, even does not exist and $\lambda = 0$, the test results are better. This shows that the classification method based on knowledge distillation used in the DTC module in [2] is redundant for our model, and our model can be trained well without the constraints of the teacher model.
  3) Test for $\gamma$
     In this part, we test for the hyper-parameter $\gamma \in [0, 1]$ as Fig.3(c) & (f), which means weight for the centrality-based distribution consistency loss term $\mathcal{L}_c^{dist}$. During our testing, we found that the variance of the F1 score is very sensitive to changes in $\gamma$, while the AUC score is only slightly affected by changes in $\gamma$. We suppose that this phenomenon caused by the $\mathcal{L}^{total}$ proposed by DTC module in [2] which does not have enough constraints on our model training. Related to our analysis of $\lambda$, we should study a new classifier suitable for HKTGNN in future work and redesign $\mathcal{L}^{total}$ to make our model perform better. $\gamma$ does not work best when it is equal to 0. The parts that are better than $\gamma = 0$ roughly include (0,0.2) and (0.5,0.8). This shows that our centrality-based loss $\mathcal{L}_c^{dist}$ is effective.
- Ablation Test
  In this part, we mainly compare the effect of our changes to the HKTGNN without centrality-based changes. The results are shown in Table IV. It can be intuitively seen that our model performs better in both F1 score and AUC score. This shows that our improvements to the algorithm are indeed effective.

## V. RELATED WORKS

### A. Supply Chain Risk Assessment

In addition to the machine learning methods mentioned in the introduction, Wichmann et al. proposed a natural language processing-based method for autonomously collecting supply chain information from the web[31]. Based on these methods, Baryannis et al. employed distributed resources to make data more efficient and improve the performance of these algorithms[4]. Link prediction has enriched supply chain risk assessment solution methods to some extent, allowing relevant practitioners to better extract useful information from real-world networks, such as hidden supply relationships and ingredient relationships, to improve supply chain visibility[1]. However, as stated in the statement, increasing awareness does not directly address the data scarcity problem. Our method takes this into consideration and improves on it.

*B. Graph Data Mining*

In addition to the use of graph data mining in supply chain risk assessment, Jha et al. proposed a protein-protein interaction prediction approach based on graph neural networks and attention mechanisms[32]. Derrow-Pinion et al. offered a graph neural network-based trip time prediction method[33]. Wu et al. reviewed the use of graph neural networks in natural language processing[34]. Wu et al. summarized graph neural network-based recommendation system method[35]. As can be seen, graph data mining is widely employed in a variety of sectors, and research into graph data mining methods is becoming increasingly important.

*C. The impact of OOD nodes on graph data mining*

OOD nodes are abundant in real graph dataset[2]. In the past, researchers recommended the following to alleviate data scarcity and distribution drift issues, as well as to lessen the influence of OOD nodes on message passing: Song et al. proposed an OOD benchmark designed exclusively for graphs[7]. Gui et al. proposed an out-of-distribution graph attention network (OODGAT), which explicitly simulates interactions between different types of nodes and distinguishes between in-value and out-value during feature propagation[25]. Li et al. presented the OOD-GNN, a novel non-linear graph representation decorrelation approach based on random Fourier features[36]. To minimize statistical dependency between related and unrelated graph representations, the model repeatedly improves the sample graph weights and graph encoder. All of these strategies have shown to be helpful in combating distribution drift. However, in all circumstances, OOD nodes do not always have noise[2]. Our CB-Graph mitigates the data scarcity problem by employing domain-differential message transmission based on centrality.

## VI. CONCLUSION&FUTURE WORK

This paper proposes a hierarchical graph neural network to model supply chain risk prediction. We build the concept of CB-Graph by easily accessible product-to-product connections and product-enterprise connections, avoiding the problem of low data quality caused by low visibility in supply chain risk assessment focusing on products in the supply chain rather than scattered enterprises in the supply chain. Through the hierarchical structure, we successfully converted the directed heterogeneous graph into a directed homogeneous graph, reducing the computation time and memory consumption of a single graph; through feature embedding and feature supplementation, we successfully improved the model's performance. At the same time, based on the centrality of domain difference feature supplementation and message passing, we solve the data hunger problem. We use an equation and widely test to prove that our model is better than others through experiments on a real-world dataset in both effective and fair.

As we discussed in Sec.IV, our future work should focus on providing a new classifier and optimizing our loss function to improve the effectiveness of our classification effect. Simultaneously, we should gather more real-world datasets to support the generality of our work.

ACKNOWLEDGMENTS

This work was supported by the National Key R&D Program of China(No.2022YFB3304300).